\documentclass[lettersize,journal]{IEEEtran}
\usepackage{amsmath,amsfonts}
\usepackage{algorithm}
\usepackage{array}
\usepackage[caption=false,font=normalsize,labelfont=sf,textfont=sf]{subfig}
\usepackage{textcomp}
\usepackage{stfloats}
\usepackage{url}
\usepackage{verbatim}
\usepackage{graphicx}
\usepackage{cite}
\usepackage{times}
\usepackage{soul}
\usepackage{url}
\usepackage[hidelinks]{hyperref}
\usepackage[utf8]{inputenc}
\usepackage[small]{caption}
\usepackage{graphicx}
\usepackage{amsmath}
\usepackage{amssymb}
\usepackage{amsthm}
\usepackage{booktabs}
\usepackage{algorithm}
\usepackage{algpseudocode}

\usepackage[switch]{lineno}
\usepackage{multirow}
\usepackage{hyperref}
\usepackage{cleveref}
\usepackage{indentfirst}
\usepackage{booktabs}
\usepackage{multirow}
\usepackage{rotating}
\usepackage[table,xcdraw]{xcolor}
\usepackage{adjustbox}
\usepackage{nccmath}
\usepackage{utfsym}

\hypersetup{
    colorlinks=true,
    linkcolor=red,
}

\crefname{figure}{Fig.}{Figs.}
\Crefname{figure}{Fig.}{Figs.}
\Crefname{table}{Table}{Tables}
\crefname{table}{Table}{Tables}

\begin{document}

\title{Improving Adversarial Transferability of Vision-Language Pre-training Models through Collaborative Multimodal Interaction}

\author{Jiyuan Fu, Zhaoyu Chen, \IEEEmembership{Graduate Student Member, IEEE}, Kaixun Jiang, \\ Haijing Guo, Jiafeng Wang, Shuyong Gao, Wenqiang Zhang, \IEEEmembership{Member, IEEE}
\thanks{Jiyuan Fu and Zhaoyu Chen contributed equally to this work. The corresponding author is Wenqiang Zhang.}
\thanks{
Jiyuan Fu, Haijing Guo, Jiafeng Wang, Shuyong Gao, and Wenqiang Zhang are with Shanghai Key Lab of Intelligent Information Processing, School of Computer Science, Fudan University, Shanghai, China.
Zhaoyu Chen, Kaixun Jiang, and Wenqiang Zhang are with Shanghai Engineering Research Center of AI \& Robotics, Academy for Engineering \& Technology, Fudan University, Shanghai, China, and also with Engineering Research Center of AI \& Robotics, Ministry of Education, Academy for Engineering \& Technology, Fudan University, Shanghai, China. The emails of these authors are fujy23@m.fudan.edu.cn, zhaoyuchen20@fudan.edu.cn, \{kxjiang22, hjguo22, jiafengwang21\}@m.fudan.edu.cn, \{sygao18, wqzhang\}@fudan.edu.cn.}}

\markboth{}%
{Fu \MakeLowercase{\textit{et al.}}: CMI-Attack}

\IEEEpubid{0000--0000/00\$00.00~\copyright~2021 IEEE}

\maketitle

\begin{abstract}
Despite the substantial advancements in Vision-Language Pre-training (VLP) models, their susceptibility to adversarial attacks poses a significant challenge. Existing work rarely studies the transferability of attacks on VLP models, resulting in a substantial performance gap from white-box attacks. We observe that prior work overlooks the interaction mechanisms between modalities, which plays a crucial role in understanding the intricacies of VLP models. In response, we propose a novel attack, called Collaborative Multimodal Interaction Attack (CMI-Attack),  leveraging modality interaction through embedding guidance and interaction enhancement. Specifically, attacking text at the embedding level while preserving semantics, as well as utilizing interaction image gradients to enhance constraints on perturbations of texts and images. Significantly, in the image-text retrieval task on Flickr30K dataset, CMI-Attack raises the transfer success rates from ALBEF to TCL, $\text{CLIP}_{\text{ViT}}$ and $\text{CLIP}_{\text{CNN}}$ by 8.11\%-16.75\% over state-of-the-art methods. Moreover, CMI-Attack also demonstrates superior performance in cross-task generalization scenarios. Our work addresses the underexplored realm of transfer attacks on VLP models, shedding light on the importance of modality interaction for enhanced adversarial robustness.
\end{abstract}

\begin{IEEEkeywords}
Adversarial example, vision-language pre-trained model, adversarial transferability, black-box attack
\end{IEEEkeywords}

\section{Introduction}
\IEEEPARstart{V}{ision-language} pre-training (VLP) models are built on the foundation of learning the intrinsic correlations between vision and language, significantly improving the performance of various tasks by providing superior feature representations. These models play a crucial role in offering a universal solution for multiple tasks, including image-text retrieval (ITR)~\cite{ITR,ITR1,ITR_TMM}, image captioning (IC)~\cite{imagecaption}, among other downstream tasks. However, recent studies have elucidated the vulnerability and sensitivity of VLP models to adversarial examples~\cite{coattack}. Investigating adversarial examples on VLP models is advantageous for designing more robust VLP models, ensuring the secure deployment of foundational models.

\IEEEpubidadjcol

\begin{figure}[t]
\centering
\includegraphics[scale=0.55]{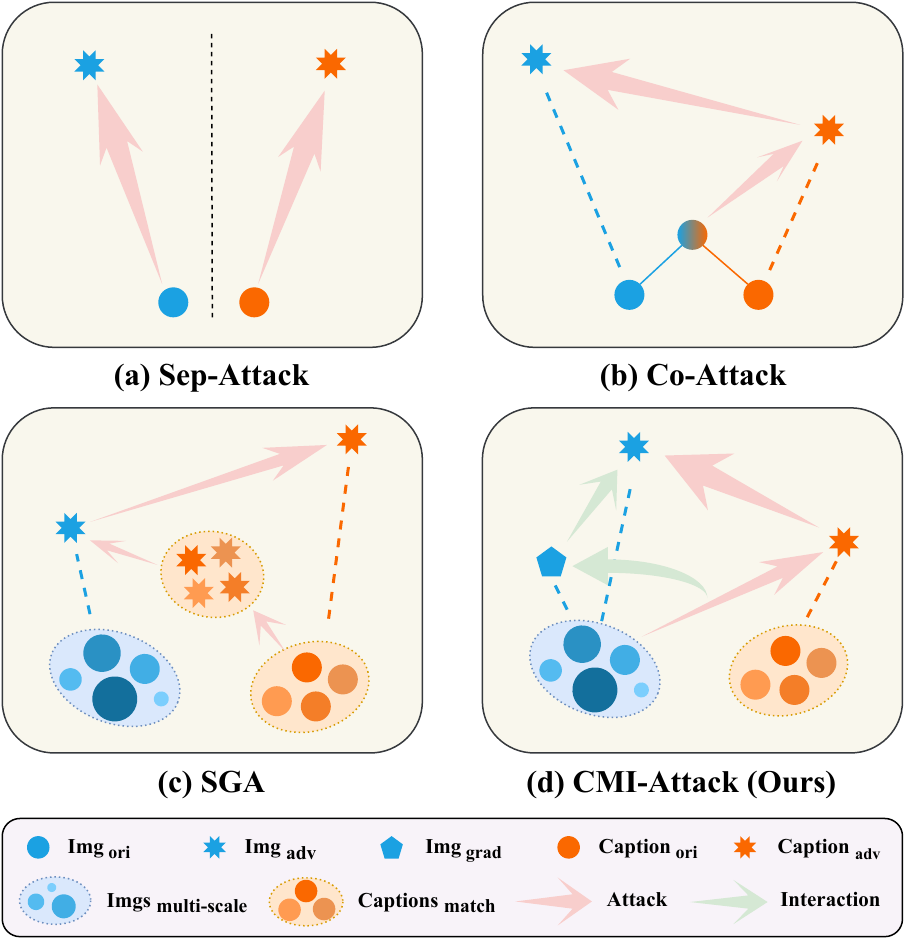} 
\caption{\textbf{Comparison of Attack Schemes for Different Attack.} Each subplot presents the attack process of the corresponding method on both images and text under specific conditions. Our approach enhances attack performance by fully leveraging the information generated through the multimodal interaction process.}
\label{Figure 1}
\end{figure}

Recently, significant progress has been made in white-box adversarial attacks on VLP models. Pioneer work shows that single-modal attack such as PGD~\cite{PGD} and BERT-Attack~\cite{bertattack} exhibit good adversarial performance in the visual and text domains. However, applying these attacks directly to VLP models still poses challenges because VLP models integrate multimodal information and require more comprehensive attack strategies to deal with their complexity. To address this issue, researchers have explored various multimodal white-box attacks, such as the Sep-Attack ~\cite{sga} that directly combines both BERT-Attack and PGD. Then Co-Attack~\cite{coattack} considers image-text collaborative information, which is specifically designed for customized attack forms for different VLP model structures. Nevertheless, white-box attacks still face limitations in practice due to a lack of access to model details such as network architecture and gradients. Therefore, black-box attacks on VLP models have emerged as a recent focal point. Black-box attacks do not require any details of the target model, which is a more realistic attack setting. By generating adversarial examples through the surrogate model, black-box attacks cause the target model to predict errors. Set-level Guidance Attack (SGA)~\cite{sga} first explores black-box attacks on VLP models and significantly improves the transferability of adversarial examples through data augmentation, many-to-many cross-modal matching, and cross-modal information guidance. However, SGA is a separate attack for each modality without fully considering the interaction between modalities. Currently, there still exists a significant gap between black-box attacks and white-box attacks, which remains an important challenge in current work.

In this paper, we focus on black-box transfer attacks against VLP models. To boost transferability, current work focuses on achieving cross-modal alignment and complementarity by attacking multiple modalities and employing multi-modal data augmentation. However, we note that existing work regrettably neglects this crucial modality interaction process, thus limiting the adversarial transferability of VLP models. In essence, \textbf{modality interaction} refers to the interaction and information exchange between different modalities. During the generation of adversarial text, direct interactive updates are applied within the same cross-modal feature space and image embedding.
Therefore, we explore the adversarial transferability of VLP models from modality interaction for the first time (in Section~\ref{sec:analysis}). Specifically, we commence with an analysis of observed phenomena in the multimodal alignment space. This includes that adversarial text generated by attacking embeddings have stronger transferability, and that gradient information from images can enhance the transferability of adversarial examples during text perturbation. In summary, these analyses unveil the importance of modality interaction, providing a clear elucidation of its significance.

Based on a profound understanding of modality interaction, we propose a novel attack on VLP models, named Collaborative Multimodal Interaction Attack (CMI-Attack), consisting of embedding guidance and interaction enhancement. First, embedding guidance utilizes similar embeddings in the multimodal feature space aligned with text and images as a guide to the highly transferable perturbing text. Then, interaction enhancement refers to fully utilizing existing alignment space information on the attack direction of images when attacking both images and text. Finally, CMI-Attack fully exploits the intricate relationships between modalities during attacks, better understanding and leveraging the mutual influence between vision and text to enhance the effectiveness of adversarial attacks. \cref{Figure 1} illustrates the distinctions and connections between our attack and previous methods. Through extensive experiments conducted on the Flickr30k~\cite{Flickr} and MSCOCO~\cite{MSCOCO} datasets, our method demonstrates stronger attack performance across different VLP models. Specifically, in the image-text retrieval task, utilizing the Flickr30k dataset, CMI-Attack achieves an 8.11\%-16.75\% increase in transfer success rates from ALBEF to TCL, $\text{CLIP}_{\text{ViT}}$, and $\text{CLIP}_{\text{CNN}}$ compared to the current state-of-the-art attacks. The main contributions of this paper can be summarized as follows:

\begin{itemize}
    \item We present the first comprehensive exploration of modality interactions during attacks on VLP models, revealing their pivotal role in bolstering adversarial attack efficacy.
    \item Our analysis unveils two pivotal discoveries: adversarial text generated from embeddings exhibits stronger transferability than word-based ones, and incorporating image gradients during text generation enhances adversarial example transferability, laying the foundation for our proposed CMI-Attack.
    \item We introduce Collaborative Multimodal Interaction Attack (CMI-Attack), a novel strategy leveraging embedding guidance and interaction enhancement, showcasing improved understanding of mutual modality influences.
    \item Through comprehensive experiments, we demonstrate that adversarial examples from CMI-Attack have stronger transferability on VLP models.
\end{itemize}

The remainder of the paper is organized as follows. Section~\ref{sec:related} briefly reviews the literature related to Vision-Language Pre-training (VLP) models, the Image-Text Retrieval (ITR) task, and adversarial transferability. Section~\ref{sec:analysis} conducts a comprehensive analysis of the transferability of existing adversarial attacks and presents two key findings on this matter. Section~\ref{sec:method} provides a detailed description of the CMI-Attack method proposed in this paper, including its specific implementation details. Section~\ref{sec:experiments} shows the experimental results to demonstrate the effectiveness of the proposed Collaborative Multimodal Interaction Attack. Finally, Section~\ref{sec:conclusions} summarizes the paper while concurrently addressing broader impacts.

\section{Related Work}
\label{sec:related}
In this section, we comprehensively review the existing work, providing an in-depth exploration of the advancements and methodologies that pave the way for our method.

\subsection{Vision-Language Pre-training Models.}
The Vision-Language Pretraining (VLP) model is a general multimodal understanding model pre-trained on a large-scale dataset of image-text pairs. This model efficiently captures the complex relationships between images and text, establishing deeper and more comprehensive visual and language representations~\cite{Multimodallearning_TMM}. From the perspective of multimodal fusion, VLP models can be classified into two categories~\cite{coattack}: Fused VLP models and Aligned VLP models. Fused VLP models ($e.g.$, ALBEF~\cite{ALBEF}, TCL~\cite{TCL}) employ a structural approach that initially utilizes a single encoder to extract feature representations from both images and text. Subsequently, a multimodal encoder performs cross-attention operations to achieve feature fusion. It is noteworthy that during the multimodal encoding process, mutual interaction and guidance between image and text features are allowed. Aligned VLP models ($e.g.$, CLIP~\cite{CLIP}) utilize a different structural approach. They exclusively use a single encoder to independently learn feature representations for images and text, and subsequently map these features into a shared semantic space. In summary, the VLP model offers a robust foundation for visual and language tasks through its multimodal fusion structure.

\subsection{Image-Text Retrieval Task.}
The primary objective of the image-text retrieval task is to establish an effective semantic association between images and text, such that when a query image or text is given, the VLP model can retrieve relevant images or text, quantifying the semantic correlation between images and text~\cite{ITR}. This task can be divided into two sub-tasks: image-to-text retrieval (TR) and text-to-image retrieval (IR). In TR, the model retrieves relevant textual information for a given image, while in IR, the model retrieves relevant images based on a given text query. In fused VLP models such as ALBEF~\cite{ALBEF} and TCL~\cite{TCL}, for each image-text pair, the semantic similarity score in the unimodal embedding feature space is first used for initial sorting, and then a multi-modal encoder is used to calculate the initial candidates to obtain the final ranking results. In contrast, in aligned VLP models such as CLIP~\cite{CLIP}, the alignment is directly based on the representation space between images and text to calculate the similarity for ranking.

\subsection{Adversarial Transferability}
Adversarial attacks are categorized into white-box attacks \cite{fgsm,chen2022shape} and black-box attacks \cite{chen2023query,chen2024content,sga} based on the extent of knowledge about the target model available to the attacker. White-box attacks refer to attackers who can access all information from the target model, such as network architecture and gradients. Black-box attacks refer to attackers who cannot obtain internal information about the target model, which is a more practical setting in the real world. 
Consequently, adversarial transferability refers to the performance and impact of attacks when transitioning from a source model to a target model.

In computer vision, white-box adversarial attacks usually are based on gradient optimization, including FGSM~\cite{fgsm}, I-FGSM~\cite{I-FGSM}, PGD~\cite{PGD}, etc. In natural language processing, common methods include TextFooler~\cite{textfooler}, BAE~\cite{BAE}, BERT-Attack~\cite{bertattack}, etc., which usually achieve the attack effect by adding, modifying, or deleting some parts of the text. In multimodal attack, Zhang \textit{et al.}\cite{coattack} combines visual and textual bimodal information and proposes the first white-box attack, Co-attack, by utilizing the synergistic effect between images and text in the VLP model. Then, SGA~\cite{sga} first explores the black-box attacks and use data augmentation to generate multiple groups of images, match them with multiple text descriptions, and comprehensively utilize cross-modal guidance information to improve the transferability of adversarial examples in black-box models. However, current attacks on VLP models, lacking consideration for modality interaction, tend to get trapped in local optima, resulting in a notable performance gap between black-box and white-box attacks.

\section{Analysis of Adversarial Transferability}
\label{sec:analysis}
While Co-Attack~\cite{coattack} and SGA~\cite{sga} achieve high attack success rates, the effectiveness of adversarial examples diminishes when transferred to other VLP models. In this section, we conduct a comprehensive analysis of the transferability of existing adversarial attacks. Here, we use the Flickr30K dataset, and the source models are uniformly set as ALBEF.

\subsection{Observation}\label{3.1}
Considering cross-modal interactions during the training of VLP models has been shown to improve their performance on downstream tasks, as indicated by existing work~\cite{cross-modal_interaction}. Motivated by this, we delve into examining the impact of modality interaction on adversarial transferability in VLP~models, conducting experiments to unveil the following two findings:

\begin{figure}[t]
\centering
\includegraphics[scale=0.23]{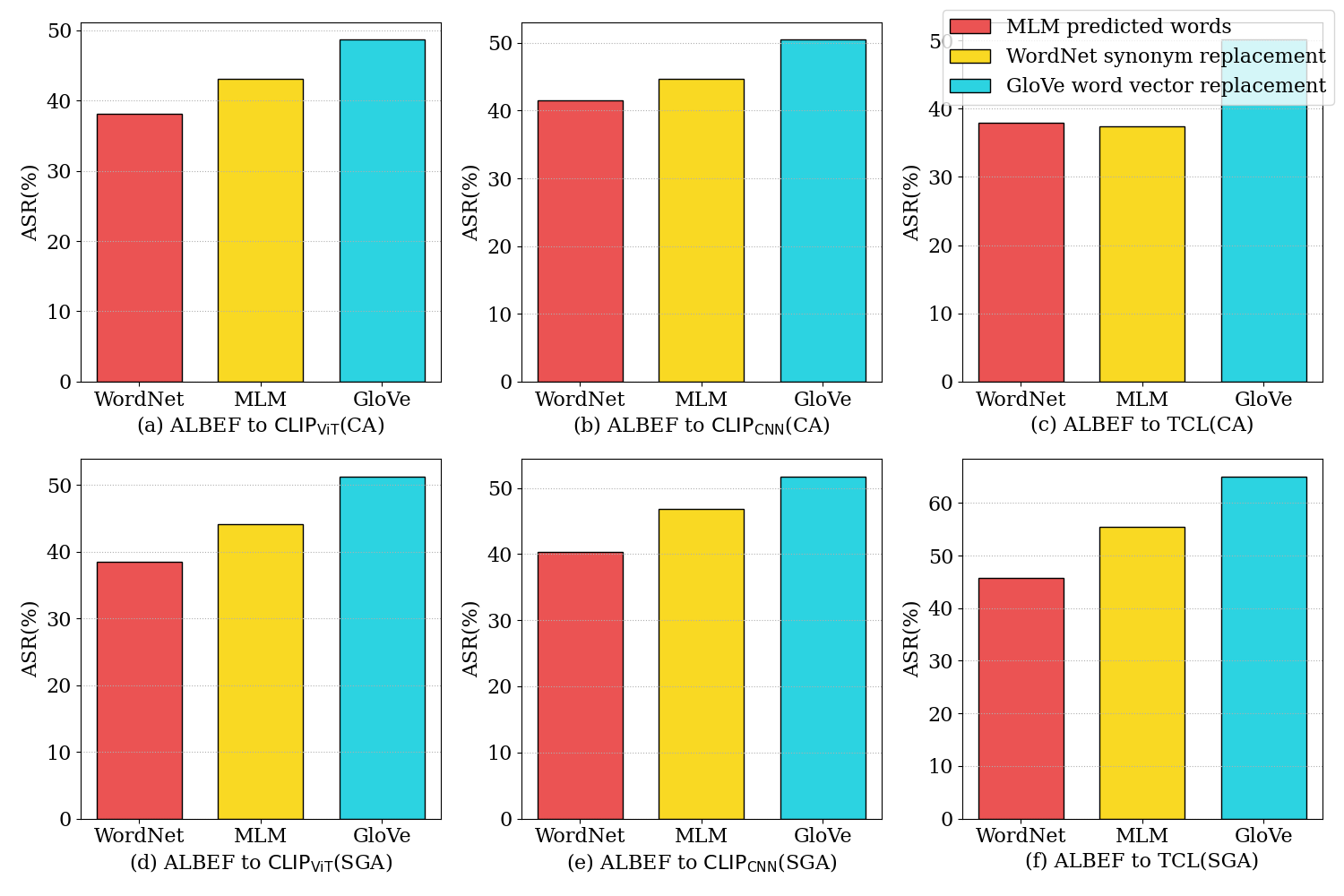} 
\caption{ \textbf{Comparing the attack success rate (ASR) on the IR R@1 metric results for image-text retrieval using different text attacks.} The two rows of bar plots display the transfer attack success rates for Co-attack and SGA, respectively. }
\label{Figure 2}
\end{figure}

\textbf{The transferability of adversarial text generated based on embeddings is stronger than that of those generated based on words.} 
The text attack in existing VLP attacks often adopts BERT-Attack, which generates adversarial texts indirectly at the word level, ignoring cross-modal correlations in the image-text alignment feature space as well as the impact of images on text generation.  To thoroughly investigate more transferable adversarial text generation methods, we conduct a series of experiments, with specific results presented in \cref{Figure 2}. For crucial words in the sentences, we apply the following perturbation settings: \textbf{i)} Performing WordNet~\cite{wordnet} synonym replacement (word level attack); \textbf{ii)} Using Masked Language Model (MLM)~\cite{BERT} for prediction (word level attack); \textbf{iii)} Employing GloVe~\cite{GloVe} word embedding for similar replacements (embedding level attack). 
Based on the observed results, it is evident that in the VLP image-text alignment space, adversarial transferability is more pronounced for perturbed text generated using embedding perubation.

\begin{figure}[t]
\centering
\includegraphics[scale=0.29]{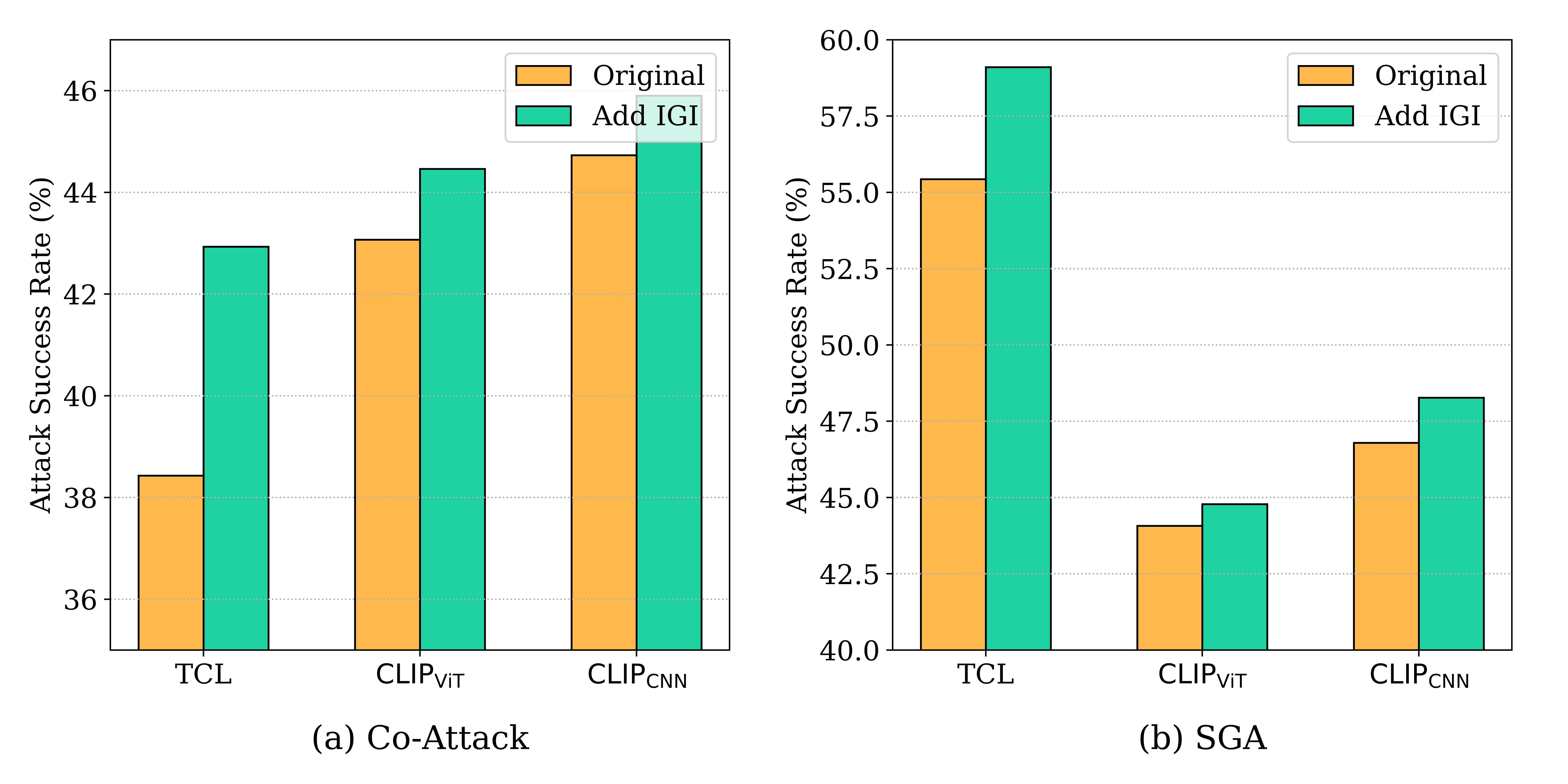} 
\caption{ \textbf{Comparing the impact of Interaction Gradient Information (IGI) on the attack success rates for the IR R@1 metric in image-text retrieval.} The first column of each bar chart represents the original method, while the second column represents the original method with interaction gradient information. }
\label{Figure 3}
\end{figure}

\textbf{The gradient information from images during the generation of adversarial text enhances the transferability of adversarial examples.} We posit that the comprehensive utilization of these gradient information enables attackers to more thoroughly consider the correlation between text and images when generating adversarial examples. Here, we define the accumulated gradients with respect to the image during the attack on the text as Interaction Gradient Information (IGI). In the subsequent sections of the paper, we refer to it simply as IGI for brevity.
In response to this hypothesis, we have implemented the following attack configurations: \textbf{i)} Employing the original Co-Attack and SGA; \textbf{ii)} Utilizing an improved attack with added Interaction Gradient Information (IGI). 
As illustrated in \cref{Figure 3}, incorporating image gradient information into the text attack not only enhances the attack success rates (ASR) of transfer attacks but also establishes a more robust and effective adversarial examples generation strategy.

\subsection{Discussion}\label{3.2}

Building upon the aforementioned observation, it becomes evident that, in the context of attacking VLP models, the absence of modal interaction processes leads to the entrapment of attack effectiveness in a local optimum. Subsequently, we delve into a comprehensive discussion regarding the limitations inherent in current VLP adversarial attacks: 

Firstly, when examining Sep-Attack, we observe that it merely combines single-modality attacks without considering the characteristics of multimodal models, leading to the limited performance of multimodal adversarial attacks. Secondly, upon closer inspection of Co-Attack, we note its reliance solely on pairwise cross-modal information, oversimplifying the potential of many-to-many cross-modal alignment. Consequently, this approach limits the generalizability of adversarial examples, particularly in black-box models. Lastly, our focus on SGA reveals that, despite its superior efficacy in leveraging multimodal information compared to Co-Attack, SGA neglects the utilization of Interaction Gradient Information (IGI), which encompasses the interactive details between text and images, in the attack process. Moreover, existing attacks conduct text attacks at the word level, lacking interaction with the image space. This limitation hinders a comprehensive understanding of the relationships between different modalities of data, impeding the efficient generation of perturbed examples.

In general, these analyzes deepen our understanding of the adversarial transferability in VLP models, and illustrate the importance of modal interactions.
Simultaneously, it further exposes the vulnerability of current VLP models in the face of adversarial examples, which is helpful to inspire researchers to design robust VLP models.

\section{Method}
\label{sec:method}
In this section, we propose an efficient and transferable attack, which we refer to as Collaborative Multimodal Interaction Attack (CMI-Attack). CMI-Attack achieves its goals by efficiently guiding attacks through the utilization of highly concealed word embeddings and enhancing transfer attack capabilities using interactive gradient information.

\subsection{Preliminaries}
Here, $i$ and $t$ denote an individual image and caption respectively. Multiple captions that best match a single image are represented as $S_t$. To enhance the transferability of adversarial examples, the original images are scaled to different sizes, denoted as $S_{i}$. Let $(i,t)$ refer to an image-text pair while $(S_{i}, S_{t})$ denotes a matching pair consisting of an image set and a caption set. Due to architectural differences in VLP models, the multimodal encoder in fused VLP models is denoted as $E_{M}$, and the text encoder in both fused and aligned VLP models is denoted as $E_{T}$, with the image encoder denoted as $E_{I}$. Specifically, $E_{I}(i)$ represents feeding an individual image into the image encoder to obtain the resulting embedding $e_{i}$. Similarly, $e_{t}$ denotes the text embedding obtained by passing a caption t into the text encoder $E_{T}(t)$. For the fused VLP models, $E_{M}(e_{i}, e_{t})$ denotes passing the image and text embeddings into the multimodal encoder $E_{M}$ to output the multimodal embedding $e_{i\&t}$. 
Moreover, we define $\epsilon_{i}$ to represent the maximum allowable perturbation for images, i.e., $||i^{adv} - i||p \leq \epsilon_{i}$, where $||\cdot||_p$ denotes the $p$-norm. Additionally, for text attacks, $\epsilon_{t}$ specifically refers to the number of perturbed words in the captions, indicating the maximum perturbation bound applicable to captions.

\subsection{Motivation}
By framing the generation of adversarial examples as an optimization problem, the transferability of adversarial examples can be regarded as a manifestation of the model's generalization ability~\cite{NISI}. Viewing the training of multimodal models, the primary objectives involve achieving modality alignment and constructing a unified vision-language alignment space. Hence, attacks should also consider these aspects. In our analysis, we observe a common issue in current adversarial attacks against VLP models, specifically the lack of corresponding updates in the multimodal feature alignment space during modality updates. Existing attack attempt to expand the distance between images and texts. However, when attacking texts and images, the guiding effect of gradients from images is relatively limited~\cite{sga}, meaning insufficient utilization of cross-modal interaction information. This oversight restricts the capability to capture the subtle changes in such multimodal matching relationships.

Therefore, we advocate employing adversarial attacks that utilize collaborative multimodal interaction to guide the perturbation process. By incorporating modal feedback to enable mutual enhancement between modalities during the attack generation, both visual and textual representations can be perturbed more effectively. This facilitates better destruction on the alignment of joint embedding spaces.

\subsection{Collaborative Multimodal Interaction Attack}
The Collaborative Multimodal Interaction Attack consists of Embedding Guidance and Interaction Enhancement. Next, we provide a detailed introduction to CMI-Attack:

\noindent\textbf{Embedding Guidance.} In Section \ref{3.2}, we analyze the text attack component in current methods targeting VLP models, revealing that the current generation of adversarial text occurs at the word level. This represents an indirect perturbation method that does not account for the influence of images on text. This limitation significantly hinders the generalization of adversarial examples across different VLP models, resulting in limited transferability of the attack.

To enhance the transfer attack success rates across different VLP models, we focus on GloVe~\cite{GloVe} word embeddings. Specifically, we utilize words with similar embeddings for text perturbation, aiming to preserve semantic similarity between the attacked text and the original text in the semantic space. The process is outlined below: 
\begin{equation}    
    S = 
    \begin{cases}
    w' \mid \text{sim}(w',w) > \tau, w' \in V_{glove}, & \text{if } w\in V_{glove}, \\
    \operatorname{argmax}_k f_{mlm}(x_t), & \text{otherwise.}
    \end{cases}
\end{equation}

For certain texts, there are no matching vectors in GloVe. Therefore, during the attack process, if the target word for the attack is not present in GloVe, we will use MLM rules for predicting word replacement attacks. Here, $w$ represents a word, $x$ denotes a sentence, $S$ denotes substitutes, $k$ denotes the number of substitute words to retain. $f_{mlm}$ represent MLM model, the process involves masking the target word, providing the masked input to the MLM model, and predicting the most probable substitute word for the masked position. Next, choose the final substitute word:

\begin{equation}
    w^* = \operatorname{argmax}_{S} \mathcal{L}(E_{T}(x_{w'}), e_i) ~, 
\end{equation}
herein, $e_i$ stands for image embedding, while the derivation of text embedding can be accomplished through $E_{T}(x_{w'})$. Ultimately, we obtain the substitute word $w^*$ by maximizing the loss function $\mathcal{L}$, a process designed to maximize the distance between the image and text modalities. 

\begin{algorithm}[t]
  \caption{Collaborative Multimodal Interaction Attack}
  \label{alg:cmi}
  \begin{algorithmic}[1]
    \Require
      Image $i$, Caption $t$, Multi-scale Image set $S_{i}=\{i_{1}, i_{2}, \dots, i_{m}\}$, Caption set $S_{t}=\{t_{1}, t_{2}, \dots, t_{m}\}$, Image encoder $E_{I}$, Text encoder $E_{T}$, iteration steps $M$, interactively enhanced iteration steps $N$, joint gradient parameters $\lambda$, Adversarial step size $\alpha$ and loss function $J$, maximum perturbation boundary of the image $\epsilon_{i}$ and the caption $\epsilon_{t}$
    \State \textit{\textbf{$//$ Generate adversarial caption set $t^{adv}$}}
    \State $g_1 \leftarrow 0$
    \For{$k\;= 1,\dots,N$}
        \State $t_{1:m}^{adv} \leftarrow \underset{t^{adv}_{1:m} \in [t_{1:m} \pm \epsilon_{t}]}{\operatorname{argmax}} - \frac{E_T(t_{1:m}^{adv})E_I(i_k)}{\|E_T(t_{1:m}^{adv})\|\|E_I(i_k)\|}$ 
        \State $g_{k+1} \leftarrow \lambda \cdot g_k + \sum_{j=1}^m \frac{\nabla \mathcal{L}(E_T(t_j^{adv}), E_I(i_j))}{\|\nabla \mathcal{L}(E_T(t_j^{adv}), E_I(i_j)) \|}$ 
        \State $i^{adv}_{k+1} \leftarrow \text{Clip}_{i,\epsilon_{i}}(i^{adv}_{k} + \alpha \cdot \text{sign}(g_{k+1})) $
    \EndFor
    \State $t^{adv} \leftarrow t^{adv}_{1:m}$
    \State \textit{\textbf{$//$ Generate adversarial image $i^{adv}$}}
    \State $g^{IGI}_1 \leftarrow g_{N+1}$
    \State $i^{adv} \leftarrow i$
    \For {$k\;= 1,\dots,M$}
        \State $g^{IGI}_{k+1} \leftarrow \lambda \cdot g^{IGI}_k + \frac{\nabla \mathcal{L}(E_T(S_t^{adv}), E_I(S_i))}{\|\nabla \mathcal{L}(E_T(S_t^{adv}), E_I(S_i)) \|}$ 
        \State $i^{adv}_{k+1} \leftarrow \text{Clip}_{i,\epsilon_{i}}(i^{adv}_{k} + \alpha \cdot \text{sign}(g^{IGI}_{k+1})) $
    \EndFor
    \State $i^{adv} \leftarrow i^{adv}_{M+1}$
    \Ensure Adversarial image $i^{adv}$, Adversarial caption $t^{adv}$.
    \end{algorithmic}

\end{algorithm}

\noindent\textbf{Interaction Enhancement.}
In the realm of VLP models, the training objective is to align information between images and text, establishing a unified alignment space. When attacking VLP models, deploying attacks within this alignment space becomes crucial.
Through a detailed analysis in~Section \ref{3.1}, we have identified that fully leveraging interaction gradient information (IGI) can enhance the transferability of adversarial examples. However, this aspect is commonly overlooked in current methods.\par
In text attacks, we perform attacks through a multimodal interaction enhancement mechanism that takes full advantage of image information in the multimodal alignment space. Specifically, first, we use image information as a constraint to perturb text. This process can be described as:
\begin{equation}
    S_t^{adv} \leftarrow \underset{S_t^{adv} \in [S_t \pm \epsilon_{t}]}{\operatorname{argmax}} - \frac{E_T(S_t^{adv})E_I(i)}{\|E_T(S_t^{adv})\|\|E_I(i)\|},
\end{equation}
aiming to encourage caption set $S_{t}=\{t_{1}, t_{2}, \dots, t_{m}\}$ to be as distant as possible from the image $i$ in the image-text alignment space. Subsequently, we incorporate the accumulation of gradients from images into the text attack process:
\begin{equation}
    g_{N+1} \leftarrow \lambda \cdot g_{N} + \sum_{j=1}^m \frac{\nabla \mathcal{L}(E_T(t_j^{adv}), E_I(i_j))}{\|\nabla \mathcal{L}(E_T(t_j^{adv}), E_I(i_j)) \|},
\end{equation}
where $(i_j, t_j)$ represents all image-text matching pairs in $(S_{i}, S_{t})$, and 
$N$ refers to the iteration count. In the first step of our method, during text attacks, the unique aspect lies in the accumulation of image gradients. This distinctive operation takes advantage of the multimodal alignment space, where image information is used as a constraint to perturb text. By accumulating gradients from images into the text attack process, we aim to constrain the perturbation of text and subsequent images, leveraging the accumulated image gradient information for enhanced adversarial examples generation.

In image attacks, we first obtain accumulated image gradient information from text attacks:
\begin{equation}
    g^{IGI} \leftarrow g_{N+1}.
\end{equation}
Next, we attack the image in the image-text alignment space to make the image set $S_i$ as distant as possible from the perturbed caption set $S_t^{adv}$. 
The detailed process is as follows:
\begin{equation}
\small
    i^{adv} \leftarrow \text{Clip}_{i,\epsilon_{i}}(i^{adv} + \alpha \cdot \text{sign}(\lambda \cdot g^{IGI} + \frac{\nabla \mathcal{L}(E_T(S_t^{adv}), E_I(S_i))}{\|\nabla \mathcal{L}(E_T(S_t^{adv}), E_I(S_i)) \|})).
\end{equation}

The detailed procedural description of the CMI-Attack can be found in Algorithm~\ref{alg:cmi}.

\begin{table*}[ht]
\centering
\scalebox{1}{
\begin{tabular}{@{}llcccccccc@{}}
\toprule
\multicolumn{10}{c}{\textbf{Flickr30K Dataset}}                                                                                                                                                                                                                                                              \\ \midrule
\multicolumn{1}{c|}{\multirow{2}{*}{\textbf{~Source Model~}}}   & \multicolumn{1}{c|}{\textbf{Target Model}}  & \multicolumn{2}{c|}{\textbf{ALBEF}}                  & \multicolumn{2}{c|}{\textbf{TCL}}                    & \multicolumn{2}{c|}{\textbf{CLIP\textsubscript{\textbf{ViT}}}}               & \multicolumn{2}{c}{\textbf{CLIP\textsubscript{\textbf{CNN}}}} \\ \cmidrule(l){2-10} 
\multicolumn{1}{c|}{}                                   & \multicolumn{1}{c|}{\textbf{Attack}}  & ~TR R@1~         & \multicolumn{1}{c|}{~IR R@1~}         & ~TR R@1~         & \multicolumn{1}{c|}{~IR R@1~}         & ~TR R@1~         & \multicolumn{1}{c|}{~IR R@1~}         & ~TR R@1~            & ~IR R@1~            \\ \midrule
\multicolumn{1}{l|}{\multirow{6}{*}{\textbf{ALBEF}}}    & \multicolumn{1}{l|}{PGD}              & 52.45$^*$          & \multicolumn{1}{c|}{58.65$^*$}          & 3.06           & \multicolumn{1}{c|}{6.79}           & 8.96           & \multicolumn{1}{c|}{13.21}          & 10.34             & 14.65             \\
\multicolumn{1}{l|}{}                                   & \multicolumn{1}{l|}{BERT-Attack}      & 11.57$^*$          & \multicolumn{1}{c|}{27.46$^*$}          & 12.64          & \multicolumn{1}{c|}{28.07}          & 29.33          & \multicolumn{1}{c|}{43.17}          & 32.69             & 46.11             \\
\multicolumn{1}{l|}{}                                   & \multicolumn{1}{l|}{Sep-Attack}       & 65.69$^*$          & \multicolumn{1}{c|}{73.95$^*$}          & 17.60          & \multicolumn{1}{c|}{32.95}          & 31.17          & \multicolumn{1}{c|}{\underline{45.23}}          & 32.83             & 45.49             \\
\multicolumn{1}{l|}{}                                   & \multicolumn{1}{l|}{Co-Attack}        & 77.16$^*$          & \multicolumn{1}{c|}{83.86$^*$}          & 15.21          & \multicolumn{1}{c|}{29.49}          & 23.60          & \multicolumn{1}{c|}{36.48}          & 25.12             & 38.89             \\
\multicolumn{1}{l|}{}                                   & \multicolumn{1}{l|}{SGA}              & \textbf{97.24$^*$} & \multicolumn{1}{c|}{\underline{97.28$^*$}}          & 45.42          & \multicolumn{1}{c|}{55.25}          & 33.38          & \multicolumn{1}{c|}{44.16}          & 34.93             & 46.57             \\
\multicolumn{1}{l|}{}                                   & \multicolumn{1}{l|}{SGA+MI}              & 96.14$^*$ & \multicolumn{1}{c|}{96.40$^*$}          & \underline{49.00}          & \multicolumn{1}{c|}{\underline{58.67}}          & \underline{33.99}          & \multicolumn{1}{c|}{44.68}          & \underline{37.29}             & \underline{47.65}             \\
\multicolumn{1}{l|}{}                                   & \multicolumn{1}{l|}{CMI-Attack~(Ours)~~~~} & \underline{97.08$^*$}          & \multicolumn{1}{c|}{\textbf{97.43$^*$}} & \textbf{62.17} & \multicolumn{1}{c|}{\textbf{69.64}} & \textbf{41.84} & \multicolumn{1}{c|}{\textbf{54.16}} & \textbf{46.23}    & \textbf{54.68}    \\ \midrule
\multicolumn{1}{l|}{\multirow{6}{*}{\textbf{TCL}}}      & \multicolumn{1}{l|}{PGD}              & 6.15           & \multicolumn{1}{c|}{10.78}          & 77.87$^*$          & \multicolumn{1}{c|}{79.48$^*$}          & 7.48           & \multicolumn{1}{c|}{13.72}          & 10.34             & 15.33             \\
\multicolumn{1}{l|}{}                                   & \multicolumn{1}{l|}{BERT-Attack}      & 11.89          & \multicolumn{1}{c|}{26.82}          & 14.54$^*$          & \multicolumn{1}{c|}{29.17$^*$}          & 29.69          & \multicolumn{1}{c|}{44.49}          & 33.46             & 46.06             \\
\multicolumn{1}{l|}{}                                   & \multicolumn{1}{l|}{Sep-Attack}       & 20.13          & \multicolumn{1}{c|}{36.48}          & 84.72$^*$          & \multicolumn{1}{c|}{86.07$^*$}          & 31.29          & \multicolumn{1}{c|}{44.65}          & 33.33             & 45.80             \\
\multicolumn{1}{l|}{}                                   & \multicolumn{1}{l|}{Co-Attack}        & 23.15          & \multicolumn{1}{c|}{40.04}          & 77.94$^*$          & \multicolumn{1}{c|}{85.59$^*$}          & 27.85          & \multicolumn{1}{c|}{41.19}          & 30.74             & 44.11             \\
\multicolumn{1}{l|}{}                                   & \multicolumn{1}{l|}{SGA}              & 48.91          & \multicolumn{1}{c|}{60.34}          & \textbf{98.37$^*$} & \multicolumn{1}{c|}{\textbf{98.81$^*$}} & 33.87          & \multicolumn{1}{c|}{44.88}          & 37.74             & 48.30             \\
\multicolumn{1}{l|}{}                                   & \multicolumn{1}{l|}{SGA+MI}              & \underline{53.60}          & \multicolumn{1}{c|}{\underline{63.52}}          & 97.26$^*$ & \multicolumn{1}{c|}{97.98$^*$} & \underline{35.83}          & \multicolumn{1}{c|}{\underline{46.10}}          & \underline{39.97}             & \underline{49.40}             \\
\multicolumn{1}{l|}{}                                   & \multicolumn{1}{l|}{CMI-Attack~(Ours)} & \textbf{61.52} & \multicolumn{1}{c|}{\textbf{71.73}} & \underline{98.00$^*$}          & \multicolumn{1}{c|}{\underline{98.67$^*$}}          & \textbf{42.45} & \multicolumn{1}{c|}{\textbf{55.06}} & \textbf{48.02}    & \textbf{56.09}    \\ \midrule
\multicolumn{1}{l|}{\multirow{6}{*}{\textbf{CLIP\textsubscript{\textbf{ViT}}}}} & \multicolumn{1}{l|}{PGD}              & 2.50           & \multicolumn{1}{c|}{4.93}           & 4.85           & \multicolumn{1}{c|}{8.17}           & 70.92$^*$          & \multicolumn{1}{c|}{78.61$^*$}          & 5.36              & 8.44              \\
\multicolumn{1}{l|}{}                                   & \multicolumn{1}{l|}{BERT-Attack}      & 9.59           & \multicolumn{1}{c|}{22.64}          & 11.80          & \multicolumn{1}{c|}{25.07}          & 28.34$^*$          & \multicolumn{1}{c|}{39.08$^*$}          & 30.40             & 37.43             \\
\multicolumn{1}{l|}{}                                   & \multicolumn{1}{l|}{Sep-Attack}       & 9.59           & \multicolumn{1}{c|}{23.25}          & 11.38          & \multicolumn{1}{c|}{25.60}          & 79.75$^*$          & \multicolumn{1}{c|}{86.79$^*$}          & 30.78             & 39.76             \\
\multicolumn{1}{l|}{}                                   & \multicolumn{1}{l|}{Co-Attack}        & 10.57          & \multicolumn{1}{c|}{24.33}          & 11.94          & \multicolumn{1}{c|}{26.69}          & 93.25$^*$          & \multicolumn{1}{c|}{95.86$^*$}          & 32.52             & 41.82             \\
\multicolumn{1}{l|}{}                                   & \multicolumn{1}{l|}{SGA}              & 13.40          & \multicolumn{1}{c|}{27.22}          & 16.23          & \multicolumn{1}{c|}{30.76}          & 99.08$^*$ & \multicolumn{1}{c|}{98.94$^*$} & 38.76             & 47.79             \\
\multicolumn{1}{l|}{}                                   & \multicolumn{1}{l|}{SGA+MI}              & \underline{13.87}          & \multicolumn{1}{c|}{\underline{28.70}}          & \underline{16.29}          & \multicolumn{1}{c|}{\underline{31.02}}          & \textbf{99.26$^*$} & \multicolumn{1}{c|}{\textbf{99.10$^*$}} & \underline{43.68}             & \underline{49.67}             \\
\multicolumn{1}{l|}{}                                   & \multicolumn{1}{l|}{CMI-Attack~(Ours)} & \textbf{18.67} & \multicolumn{1}{c|}{\textbf{37.58}} & \textbf{21.18} & \multicolumn{1}{c|}{\textbf{39.83}} & \underline{98.16}$^*$          & \multicolumn{1}{c|}{\underline{98.81$^*$}}          & \textbf{49.04}    & \textbf{56.71}    \\ \midrule
\multicolumn{1}{l|}{\multirow{6}{*}{\textbf{CLIP\textsubscript{\textbf{CNN}}}}} & \multicolumn{1}{l|}{PGD}              & 2.09           & \multicolumn{1}{c|}{4.82}           & 4.00           & \multicolumn{1}{c|}{7.81}           & 1.10           & \multicolumn{1}{c|}{6.60}           & 86.46$^*$             & 92.25$^*$             \\
\multicolumn{1}{l|}{}                                   & \multicolumn{1}{l|}{BERT-Attack}      & 8.86           & \multicolumn{1}{c|}{23.27}          & 12.33          & \multicolumn{1}{c|}{25.48}          & 27.12          & \multicolumn{1}{c|}{37.44}          & 30.40$^*$             & 40.10$^*$             \\
\multicolumn{1}{l|}{}                                   & \multicolumn{1}{l|}{Sep-Attack}       & 8.55           & \multicolumn{1}{c|}{23.41}          & 12.64          & \multicolumn{1}{c|}{26.12}          & 28.34          & \multicolumn{1}{c|}{39.43}          & 91.44$^*$             & 95.44$^*$             \\
\multicolumn{1}{l|}{}                                   & \multicolumn{1}{l|}{Co-Attack}        & 8.79           & \multicolumn{1}{c|}{23.74}          & 13.10          & \multicolumn{1}{c|}{26.07}          & 28.79          & \multicolumn{1}{c|}{40.03}          & 94.76$^*$             & 96.89$^*$             \\
\multicolumn{1}{l|}{}                                   & \multicolumn{1}{l|}{SGA}              & 11.42          & \multicolumn{1}{c|}{24.80}          & 14.91          & \multicolumn{1}{c|}{28.82}          & 31.24          & \multicolumn{1}{c|}{42.12}          & 99.24$^*$             & 99.49$^*$             \\
\multicolumn{1}{l|}{}                                   & \multicolumn{1}{l|}{SGA+MI}              & \underline{12.30}          & \multicolumn{1}{c|}{\underline{25.40}}          & \underline{15.31}          & \multicolumn{1}{c|}{\underline{29.58}}          & \underline{31.29}          & \multicolumn{1}{c|}{\underline{42.69}}          & \underline{99.62$^*$}             & \underline{99.74$^*$}             \\
\multicolumn{1}{l|}{}                                   & \multicolumn{1}{l|}{CMI-Attack~(Ours)} & \textbf{18.87} & \multicolumn{1}{c|}{\textbf{37.98}} & \textbf{22.23} & \multicolumn{1}{c|}{\textbf{40.07}} & \textbf{39.39} & \multicolumn{1}{c|}{\textbf{52.29}} & \textbf{99.74$^*$}    & \textbf{99.76$^*$}    \\ \bottomrule
\end{tabular}
}
\caption{\textbf{Attack success rate (\%) in image-text retrieval on the Flickr30K dataset.} We present the attack success rate metric R@1 for both IR and TR tasks, indicating the success of attacks at Rank 1. $^*$ indicates white-box attacks.}
\label{Table 1}
\end{table*}

\begin{table*}[ht]
\centering
\scalebox{1}{
\begin{tabular}{@{}llcccccccc@{}}
\toprule
\multicolumn{10}{c}{\textbf{MSCOCO Dataset}}                                                                                                                                                                                                                                                \\ \midrule
\multicolumn{1}{c|}{\multirow{2}{*}{\textbf{~Source Model~}}}   & \multicolumn{1}{c|}{\textbf{Target Model}}      & \multicolumn{2}{c|}{\textbf{ALBEF}}                           & \multicolumn{2}{c|}{\textbf{TCL}}                             & \multicolumn{2}{c|}{\textbf{CLIP\textsubscript{\textbf{ViT}}}}                        & \multicolumn{2}{c}{\textbf{CLIP\textsubscript{\textbf{CNN}}}}    \\ \cmidrule(l){2-10} 
\multicolumn{1}{c|}{}                          & \multicolumn{1}{c|}{\textbf{Attack}}      & ~TR R@1~         & \multicolumn{1}{c|}{~IR R@1~}         & ~TR R@1~         & \multicolumn{1}{c|}{~IR R@1~}         & ~TR R@1~         & \multicolumn{1}{c|}{~IR R@1~}         & ~TR R@1~         & ~IR R@1~         \\ \midrule
\multicolumn{1}{l|}{\multirow{6}{*}{\textbf{ALBEF}}}    & \multicolumn{1}{l|}{PGD}         & 76.70$^*$          & \multicolumn{1}{c|}{86.30$^*$}          & 12.46          & \multicolumn{1}{c|}{17.77}          & 13.96          & \multicolumn{1}{c|}{23.10}          & 17.45          & 23.54          \\
\multicolumn{1}{l|}{}                          & \multicolumn{1}{l|}{BERT-Attack} & 24.39$^*$          & \multicolumn{1}{c|}{36.13$^*$}          & 24.34          & \multicolumn{1}{c|}{33.39}          & 44.94          & \multicolumn{1}{c|}{52.28}          & 47.73          & 54.75          \\
\multicolumn{1}{l|}{}                          & \multicolumn{1}{l|}{Sep-Attack}  & 82.60$^*$          & \multicolumn{1}{c|}{89.88$^*$}          & 32.83          & \multicolumn{1}{c|}{42.92}          & 44.03          & \multicolumn{1}{c|}{54.46}          & 46.96          & 55.88          \\
\multicolumn{1}{l|}{}                          & \multicolumn{1}{l|}{Co-Attack}   & 79.87$^*$          & \multicolumn{1}{c|}{87.83$^*$}          & 32.62          & \multicolumn{1}{c|}{43.09}          & 44.89          & \multicolumn{1}{c|}{54.75}          & 47.30          & 55.64          \\
\multicolumn{1}{l|}{}                          & \multicolumn{1}{l|}{SGA}         & \underline{96.75$^*$}          & \multicolumn{1}{c|}{\underline{96.95$^*$}}          & 58.56          & \multicolumn{1}{c|}{65.38}          & 57.06          & \multicolumn{1}{c|}{65.25}          & 58.95          & 66.52          \\
\multicolumn{1}{l|}{}                          & \multicolumn{1}{l|}{SGA+MI}         & 96.03$^*$          & \multicolumn{1}{c|}{96.26$^*$}          & \underline{60.82}          & \multicolumn{1}{c|}{\underline{67.02}}          & \underline{59.69}          & \multicolumn{1}{c|}{\underline{65.88}}          & \underline{59.79}          & \underline{67.54}          \\
\multicolumn{1}{l|}{}                          & \multicolumn{1}{l|}{CMI-Attack~(Ours)~~~~}        & \textbf{97.40$^*$} & \multicolumn{1}{c|}{\textbf{97.51$^*$}} & \textbf{72.09} & \multicolumn{1}{c|}{\textbf{75.57}} & \textbf{64.56} & \multicolumn{1}{c|}{\textbf{71.12}} & \textbf{66.37} & \textbf{72.59} \\ \midrule
\multicolumn{1}{l|}{\multirow{6}{*}{\textbf{TCL}}}      & \multicolumn{1}{l|}{PGD}         & 10.83          & \multicolumn{1}{c|}{16.52}          & 59.58$^*$          & \multicolumn{1}{c|}{69.53$^*$}          & 14.23          & \multicolumn{1}{c|}{22.28}          & 17.25          & 23.12          \\
\multicolumn{1}{l|}{}                          & \multicolumn{1}{l|}{BERT-Attack} & 35.32          & \multicolumn{1}{c|}{45.92}          & 38.54$^*$          & \multicolumn{1}{c|}{48.48$^*$}          & 51.09          & \multicolumn{1}{c|}{58.80}          & 52.23          & 61.26          \\
\multicolumn{1}{l|}{}                          & \multicolumn{1}{l|}{Sep-Attack}  & 41.71          & \multicolumn{1}{c|}{52.97}          & 70.32$^*$          & \multicolumn{1}{c|}{78.97$^*$}          & 50.74          & \multicolumn{1}{c|}{60.13}          & 51.90          & 61.26          \\
\multicolumn{1}{l|}{}                          & \multicolumn{1}{l|}{Co-Attack}   & 46.08          & \multicolumn{1}{c|}{57.09}          & 85.38$^*$          & \multicolumn{1}{c|}{91.39$^*$}          & 51.62          & \multicolumn{1}{c|}{60.46}          & 52.13          & 62.49          \\
\multicolumn{1}{l|}{}                          & \multicolumn{1}{l|}{SGA}         & 65.93          & \multicolumn{1}{c|}{73.30}           & \textbf{98.97$^*$} & \multicolumn{1}{c|}{\underline{99.15$^*$}}          & 56.34          & \multicolumn{1}{c|}{63.99}          & 59.44          & 65.70           \\
\multicolumn{1}{l|}{}                          & \multicolumn{1}{l|}{SGA+MI}         & \underline{68.68}          & \multicolumn{1}{c|}{\underline{76.08}}           & 98.89$^*$ & \multicolumn{1}{c|}{98.87$^*$}          & \underline{57.69}          & \multicolumn{1}{c|}{\underline{65.00}}          & \underline{59.99}          & \underline{67.12}           \\
\multicolumn{1}{l|}{}                          & \multicolumn{1}{l|}{CMI-Attack~(Ours)}        & \textbf{78.63} & \multicolumn{1}{c|}{\textbf{82.55}} & \underline{98.94$^*$}          & \multicolumn{1}{c|}{\textbf{99.30$^*$}} & \textbf{66.65} & \multicolumn{1}{c|}{\textbf{71.91}} & \textbf{68.66} & \textbf{74.01} \\ \midrule
\multicolumn{1}{l|}{\multirow{6}{*}{\textbf{CLIP\textsubscript{\textbf{ViT}}}}} & \multicolumn{1}{l|}{PGD}         & 7.24           & \multicolumn{1}{c|}{10.75}          & 10.19          & \multicolumn{1}{c|}{13.74}          & 54.79$^*$          & \multicolumn{1}{c|}{66.85$^*$}          & 7.32           & 11.34          \\
\multicolumn{1}{l|}{}                          & \multicolumn{1}{l|}{BERT-Attack} & 20.34          & \multicolumn{1}{c|}{29.74}          & 21.08          & \multicolumn{1}{c|}{29.61}          & 45.06$^*$          & \multicolumn{1}{c|}{51.68$^*$}          & 44.54          & 55.32          \\
\multicolumn{1}{l|}{}                          & \multicolumn{1}{l|}{Sep-Attack}  & 23.41          & \multicolumn{1}{c|}{34.61}          & 25.77          & \multicolumn{1}{c|}{36.84}          & 68.52$^*$          & \multicolumn{1}{c|}{77.94$^*$}          & 43.11          & 49.76          \\
\multicolumn{1}{l|}{}                          & \multicolumn{1}{l|}{Co-Attack}   & 30.28          & \multicolumn{1}{c|}{42.67}          & 32.84          & \multicolumn{1}{c|}{44.69}          & 97.98$^*$          & \multicolumn{1}{c|}{98.80$^*$}          & 55.08          & 62.51          \\
\multicolumn{1}{l|}{}                          & \multicolumn{1}{l|}{SGA}         & 33.41          & \multicolumn{1}{c|}{44.64}          & \underline{37.54}          & \multicolumn{1}{c|}{\underline{47.76}}          & \textbf{99.79$^*$} & \multicolumn{1}{c|}{\textbf{99.79$^*$}} & 58.93          & 65.83          \\
\multicolumn{1}{l|}{}                          & \multicolumn{1}{l|}{SGA+MI}         & \underline{34.93}          & \multicolumn{1}{c|}{\underline{45.70}}          & 35.87          & \multicolumn{1}{c|}{46.68}          & \underline{99.77$^*$} & \multicolumn{1}{c|}{98.67$^*$} & \underline{61.26}          & \underline{67.84}          \\
\multicolumn{1}{l|}{}                          & \multicolumn{1}{l|}{CMI-Attack~(Ours)}        & \textbf{48.41} & \multicolumn{1}{c|}{\textbf{58.26}} & \textbf{49.10} & \multicolumn{1}{c|}{\textbf{60.16}} & 99.47$^*$          & \multicolumn{1}{c|}{\underline{99.73$^*$}}          & \textbf{72.62} & \textbf{76.07} \\ \midrule
\multicolumn{1}{l|}{\multirow{6}{*}{\textbf{CLIP\textsubscript{\textbf{CNN}}}}} & \multicolumn{1}{l|}{PGD}         & 7.01           & \multicolumn{1}{c|}{10.62}          & 10.08          & \multicolumn{1}{c|}{13.65}          & 4.88           & \multicolumn{1}{c|}{10.70}          & 76.99$^*$          & 84.20$^*$          \\
\multicolumn{1}{l|}{}                          & \multicolumn{1}{l|}{BERT-Attack} & 23.38          & \multicolumn{1}{c|}{34.64}          & 24.58          & \multicolumn{1}{c|}{29.61}          & 51.28          & \multicolumn{1}{c|}{57.49}          & 54.43$^*$          & 62.17$^*$          \\
\multicolumn{1}{l|}{}                          & \multicolumn{1}{l|}{Sep-Attack}  & 26.53          & \multicolumn{1}{c|}{39.29}          & 30.26          & \multicolumn{1}{c|}{41.51}          & 50.44          & \multicolumn{1}{c|}{57.11}          & 88.72$^*$          & 92.49$^*$          \\
\multicolumn{1}{l|}{}                          & \multicolumn{1}{l|}{Co-Attack}   & 29.83          & \multicolumn{1}{c|}{41.97}          & 32.97          & \multicolumn{1}{c|}{43.72}          & 53.10          & \multicolumn{1}{c|}{58.90}          & 96.72$^*$          & 98.56$^*$          \\
\multicolumn{1}{l|}{}                          & \multicolumn{1}{l|}{SGA}         & 31.61          & \multicolumn{1}{c|}{43.00}          & \underline{34.81}          & \multicolumn{1}{c|}{\underline{45.95}}          & 56.62          & \multicolumn{1}{c|}{60.77}          & 99.61$^*$          & 99.80$^*$           \\
\multicolumn{1}{l|}{}                          & \multicolumn{1}{l|}{SGA+MI}         & \underline{31.84}          & \multicolumn{1}{c|}{\underline{43.38}}          & 33.23          & \multicolumn{1}{c|}{44.77}          & \underline{57.99}          & \multicolumn{1}{c|}{\underline{60.98}}          & \underline{99.71$^*$}          & \underline{99.92$^*$}           \\
\multicolumn{1}{l|}{}                          & \multicolumn{1}{l|}{CMI-Attack~(Ours)}        & \textbf{45.30} & \multicolumn{1}{c|}{\textbf{55.62}} & \textbf{47.78} & \multicolumn{1}{c|}{\textbf{57.94}} & \textbf{67.72} & \multicolumn{1}{c|}{\textbf{70.83}} & \textbf{99.75$^*$} & \textbf{99.95$^*$} \\ \bottomrule
\end{tabular}
}
\caption{\textbf{Attack success rate (\%) in image-text retrieval on the MSCOCO Dataset.} We present the attack success rate metric R@1 for both IR and TR tasks, indicating the success of attacks at Rank 1. $^*$ indicates white-box attacks.}
\label{Table 2}
\end{table*}

\begin{table}[t]
\centering
\begin{adjustbox}{max width=\columnwidth}
\begin{tabular}{@{}cccccc@{}}
\toprule
\multicolumn{6}{c}{\textbf{Image Caption}}                                                                      \\ \midrule
\multicolumn{1}{c|}{Attack}    & ~B@4~           & ~METEOR~        & ~ROUGE-L~       & ~CIDEr~          & ~SPICE~         \\ \midrule
\multicolumn{1}{c|}{Baseline}  & 39.7          & 31.0            & 60.0            & 133.3          & 23.8          \\
\multicolumn{1}{c|}{Co-Attack} & 37.4          & 29.8          & 58.4          & 125.5          & 22.8          \\
\multicolumn{1}{c|}{SGA}       & 34.8          & 28.4          & 56.3          & 116.0          & 21.4          \\
\multicolumn{1}{c|}{Ours}      & \textbf{33.9} & \textbf{27.3} & \textbf{55.2} & \textbf{108.8} & \textbf{20.3} \\ \bottomrule
\end{tabular}
\end{adjustbox}
\caption{\textbf{Cross-Task Transferability: Image Captioning. }The experimental setup utilized the MSCOCO dataset with ALBEF as the source model and BLIP as the target model. The baseline reflects the original performance of Image Caption on clean data, with lower scores indicating better attack effectiveness.}
\label{Table 3}
\end{table}

\section{Experiments}
\label{sec:experiments}
In this section, we perform a comprehensive set of experiments to evaluate our approach. We begin by detailing the experimental setup, specifying datasets, models, and metrics used. Next, we evaluate CMI-Attack performance across diverse downstream tasks, conduct an ablation study on its components, and present visualization results. These experiments aim to thoroughly analyze and validate the efficacy of our proposed approach.

\subsection{Setup}
\noindent\textbf{Datasets.}
Our experimental setup involves two widely recognized datasets in the field of visual language understanding: Flickr30K~\cite{Flickr} and MSCOCO~\cite{MSCOCO}. Specifically, for the Image Retrieval (IR) and Text Retrieval (TR) tasks, the Flickr30K dataset comprises 1000 images paired with 5000 captions, while the MSCOCO dataset, renowned for its scale, provides a larger dataset with 5000 images and 25000 captions. Additionally, we use the MSCOCO dataset's val2014 subset to validate our adversarial attack for Image Caption.\par
\noindent\textbf{Models.}
We study two main types of vision-language models: fused VLP models and aligned VLP models. The fused models early fused visual and language features when processing two modal data, and achieved joint processing through shared layers. ALBEF~\cite{ALBEF} and TCL~\cite{TCL} are representative fused models, both of which use 12-layer ViT-B/16~\cite{ViT} as the image encoder, and two independent 6-layer Transformers for text and multimodal encoding. The main difference is their distinct pre-training objectives. On the contrary, the aligned model initially independently processed images and text, and then aligned multimodal representations in the deep layer to learn the complex relationship between the two. We also examine two variants of CLIP~\cite{CLIP}, namely $\text{CLIP}_{\text{ViT}}$ and $\text{CLIP}_{\text{CNN}}$. The former is based on ViT-B/16 image encoder, while the latter uses ResNet-101~\cite{ResNet} as the backbone network. By evaluating the transferability of adversarial examples between the source and the target model, we verify the effectiveness of the proposed framework.\par
\noindent\textbf{Attacks Settings.}
We compare with the previous attack methods PGD, BERT-Attack, Sep-Attack, Co-Attack, and state-of-the-art (SOTA) approach SGA in the Image-Text Retrieval (ITR) task.
In terms of image attacks, we employ the PGD~\cite{PGD} method with the image perturbation bound set to $\epsilon_{i}=2/255$, step size $\alpha=0.5/255$, iteration steps $M=10$, interactive enhanced iteration steps $N=1$, and joint gradient parameters $\lambda=0.9$. For text attack, we utilize the BERT-Attack~\cite{bertattack} method with the perturbation bound $\epsilon_{t}$ set to 1. Additionally, following the conditions set in the SGA~\cite{sga} experiments, we scale the images at five different levels (0.5, 0.75, 1.0, 1.25, and 1.5) to enrich the diversity of the image. Additionally, for a fair experimental comparison, we conduct experiments by augmenting SGA with MIFGSM~\cite{mifgsm} (SGA+MI), where the parameter $\lambda=0.9$.\par

\noindent\textbf{Metrics.}
To evaluate the effectiveness of our proposed attack, we employ the Attack Success Rate (ASR) metric to measure the capability of white-box attacks and the transferability of black-box attacks against VLP models. The ASR metric reflects the overall success rate of attacks, and a higher ASR value indicates better attack performance. We also adopt IR R@k, representing the proportion that none of the top-k image retrieval results contain the correct images. Similarly, we utilize TR R@k, denoting the proportion that none of the top-k subtitle retrieval results have the correct matching subtitle.

\begin{table}[t]
\centering
\scalebox{1}{
\begin{tabular}{@{}cc|ccc@{}}
\toprule
\multicolumn{2}{c|}{\textbf{Module}}                                                       & \multicolumn{3}{c}{\textbf{Combined R@1 Average}}                                                    \\ \midrule
~~~~EG~~                           & \multicolumn{1}{c|}{~~IE~~~}                           & ~~~TCL~~   & ~~CLIP\textsubscript{ViT}~~ & ~CLIP\textsubscript{CNN}~~~~ \\ \midrule
~~\usym{2718} & \multicolumn{1}{c|}{\usym{2718}} & ~51.38 & 41.96                                    & 44.87~~                                    \\ \midrule
~~\usym{2718} & \multicolumn{1}{c|}{\usym{2714}} & ~54.10 & 42.54                                    & 45.49~~                                    \\ \midrule
~~\usym{2714} & \multicolumn{1}{c|}{\usym{2718}} & ~62.19 & 47.33                                    & 49.54~~                                    \\ \midrule
~~\usym{2714} & \multicolumn{1}{c|}{\usym{2714}} & ~65.91 & 48.00                                    & 50.46~~                                    \\ \bottomrule
\end{tabular}
}
\caption{\textbf{Impact of Embedding Guidance (EG) and Interaction Enhancement (IE) Modules on Adversarial Transferability.} We use the Flickr30K dataset with ALBEF as the source model. The "Combined R@1 Average" represents the average of IR R@1 and TR R@1. ASR (\%) assesses the transferability of adversarial attacks.
}
\label{Table 7}
\vspace{-8pt}
\end{table}

\subsection{Transferability Comparison}
\noindent\textbf{Image-Text Retrieval.} Our experiment focuses on the task of VLP Image-Text Retrieval (ITR). We generate adversarial examples on various source models and assess the effectiveness of different methods by calculating the ASR of white-box attacks and attacks transferred to other models. As demonstrated in \cref{Table 1} and \cref{Table 2}, our approach outperforms state-of-the-art methods in terms of attack success rate in transfer attacks on both the Flickr30K~\cite{Flickr} and MSCOCO datasets~\cite{MSCOCO}. Specifically, on the Flickr30K dataset with ALBEF~\cite{ALBEF} set as the source model, the attack success rate of transfer attacks against other VLP black-box models obtains a significant improvement of 8.11\%-16.75\%. Similarly, under the TCL~\cite{TCL} source model, the attack success rate of transfer attacks increases by 7.79\%-12.61\%. In evaluations on the larger MSCOCO dataset, adversarial examples crafted from ALBEF improve the attack success rate of transfer attacks to other models by 5.87\%-13.53\%; while starting from TCL, the improvement reaches 7.92\%-12.7\%. It is worth noting that, even with the addition of MIFGSM to SGA, our method significantly outperforms it in terms of attack success rates. In essence, the substantial augmentation in black-box attack transferability for image-text retrieval, as observed in our results, can be attributed to our method's meticulous emphasis on modality interaction and the judicious utilization of information across modalities during the adversarial generation process. These strategic choices significantly contribute to the observed advancements within a more condensed framework.

\noindent\textbf{Image Caption.} In the realm of multimodal adversarial attacks, the successful cross-task transfer underscores the effectiveness of the attack methodology. The image captioning task aims to enable models to extract visual information from images and generate corresponding textual descriptions, facilitating cross-modal visual and language understanding within the model~\cite{IC_TMM,IC_TMM_2,IC_TMM_3}. In our experimental framework, we leverage the MSCOCO dataset~\cite{MSCOCO} and designate the ALBEF~\cite{ALBEF} as the source model to craft adversarial examples, which are then used to mount attacks against the BLIP~\cite{BLIP} as the target model. To offer comprehensive quantifications of attack success, we adopt five evaluation metrics: BLEU-4 (B@4)~\cite{bleu}, METEOR~\cite{METEOR}, ROUGE-L~\cite{rouge}, CIDEr~\cite{cider},~and SPICE~\cite{SPICE}. This set of metrics enables holistic assessments on the quality, diversity, and semantic consistency of the generated text, serving as standard benchmarks for tasks including image captioning. Notably, the degradation in these scores validates the intensified efficacy of our cross-task attacks. The experimental results are illustrated in \cref{Table 3}, where, compared to the current state-of-the-art methods, the CMI-Attack demonstrates a more powerful performance. Particularly noteworthy is the 7.2\% improvement observed in the CIDEr metric for CMI-Attack. This achievement is credited to our approach, which capitalizes on the modality interaction between vision and text. By significantly improving the effectiveness of adversarial attacks, our method convincingly showcases advanced attack efficacy in cross-task transfer scenarios as well.

\begin{table}[t]
\centering
\scalebox{0.96}{
\begin{tabular}{@{}c|cc|cc|cc@{}}
\toprule
\multirow{2}{*}{ \textit{$\lambda$} } & \multicolumn{2}{c|}{TCL} & \multicolumn{2}{c|}{CLIP\textsubscript{ViT}} & \multicolumn{2}{c}{CLIP\textsubscript{CNN}} \\ \cmidrule(l){2-7} 
                                       & TR R@1  & IR R@1         & TR R@1    & IR R@1           & TR R@1   & IR R@1           \\ \midrule
0                                      & 57.96   & 67.98          & 40.37     & 53.19            & 45.72    & 53.62            \\
0.5                                    & 59.54   & 69.07          & 41.47     & 53.77            & 45.47    & 53.79            \\
0.8                                    & 61.64   & 69.45          & 41.84     & 53.74            & 46.36    & 54.68            \\
0.9                                    & \textbf{62.17}   & \textbf{69.64} & \textbf{41.96}     & \textbf{54.19}   & 46.10    & 54.65            \\
1                                      & 61.96   & 69.60          & 41.10     & 54.16            & 46.10    & 54.58            \\
1.3                                    & 59.85   & 68.19          & 41.47     & 53.61            & \textbf{46.49}    & \textbf{54.99}   \\
1.8                                    & 55.85   & 65.33          & 41.72     & 54.12            & 46.10    & 54.61            \\ \bottomrule
\end{tabular}
}

\caption{\textbf{Impact of Joint Gradient $\lambda$ on Results. }
The dataset used is Flickr30K, and the source model is ALBEF. ASR (\%) are employed to quantify the transferability of adversarial examples.
}
\label{Table 4}

\end{table}

\subsection{Ablation Study}
In the ablation study, we adopt Flickr30K as the dataset and ALBEF as the source model. ASR (\%) is utilized as the metric to quantify the transferability of adversarial examples.

\noindent\textbf{Module Ablation Experiment.} 
In addition, we conduct a series of ablation experiments to assess the impact of the Embedding Guidance (EG) and Interaction Enhancement (IE) modules on our approach in the context of transfer attacks. 
As evidenced in \cref{Table 7}, the incorporation of the EG module results in a significant improvement in transfer attack capability. Specifically, using only the EG module increases the attack success rate against TCL from 51.38\% to 62.19\%. Similarly, when targeting CLIP\textsubscript{ViT}, the attack success rate rises from 41.96\% to 47.33\%. This indicates that the EG module better accounts for inter-modality influences when handling text attacks, thereby enhancing the transferability of the attack. 
Moreover, the incorporation of the IE module contributes~to performance gains; when both modules are introduced simultaneously, the black-box attack success rate against the TCL model reaches 65.91\%, demonstrating the synergistic effect between them and resulting in a more pronounced performance improvement. In summary, our ablation study provides quantitative evidence that the Collaborative Multimodal Interaction mechanism, which includes the EG and IE modules, is crucial for enhancing adversarial transferability against VLP models.

\begin{table}[t]
\fontsize{8pt}{9pt}\selectfont
\centering
\begin{adjustbox}{max width=\columnwidth}
\begin{tabular}{@{}l|cc|cc|cc@{}}
\toprule
\multirow{2}{*}{\textit{Method}} & \multicolumn{2}{c|}{TCL}        & \multicolumn{2}{c|}{CLIP\textsubscript{ViT}}    & \multicolumn{2}{c}{CLIP\textsubscript{CNN}}     \\ \cmidrule(l){2-7} 
                        & TR R@1         & IR R@1         & TR R@1         & IR R@1         & TR R@1         & IR R@1         \\ \midrule
MLM                     & 48.68          & 59.52          & 36.56          & 48.49          & 40.23          & 50.74          \\
Wordnet                 & 30.66          & 36.98          & 12.39          & 18.81          & 13.67          & 19.28          \\
Glove                   & 61.75          & 69.40          & 41.10          & 53.70          & \textbf{46.10} & 54.34          \\
Ours                    & \textbf{62.17} & \textbf{69.64} & \textbf{41.96} & \textbf{54.19} & \textbf{46.10} & \textbf{54.65} \\ \bottomrule
\end{tabular}
\end{adjustbox}
\caption{\textbf{Impact of Different Text Attack Strategies.} The experiment configuration employs the Flickr30K dataset, utilizing ALBEF as the source model.}
\label{Table 5}
\end{table}

\begin{figure}[t]
\centering
\includegraphics[scale=0.4]{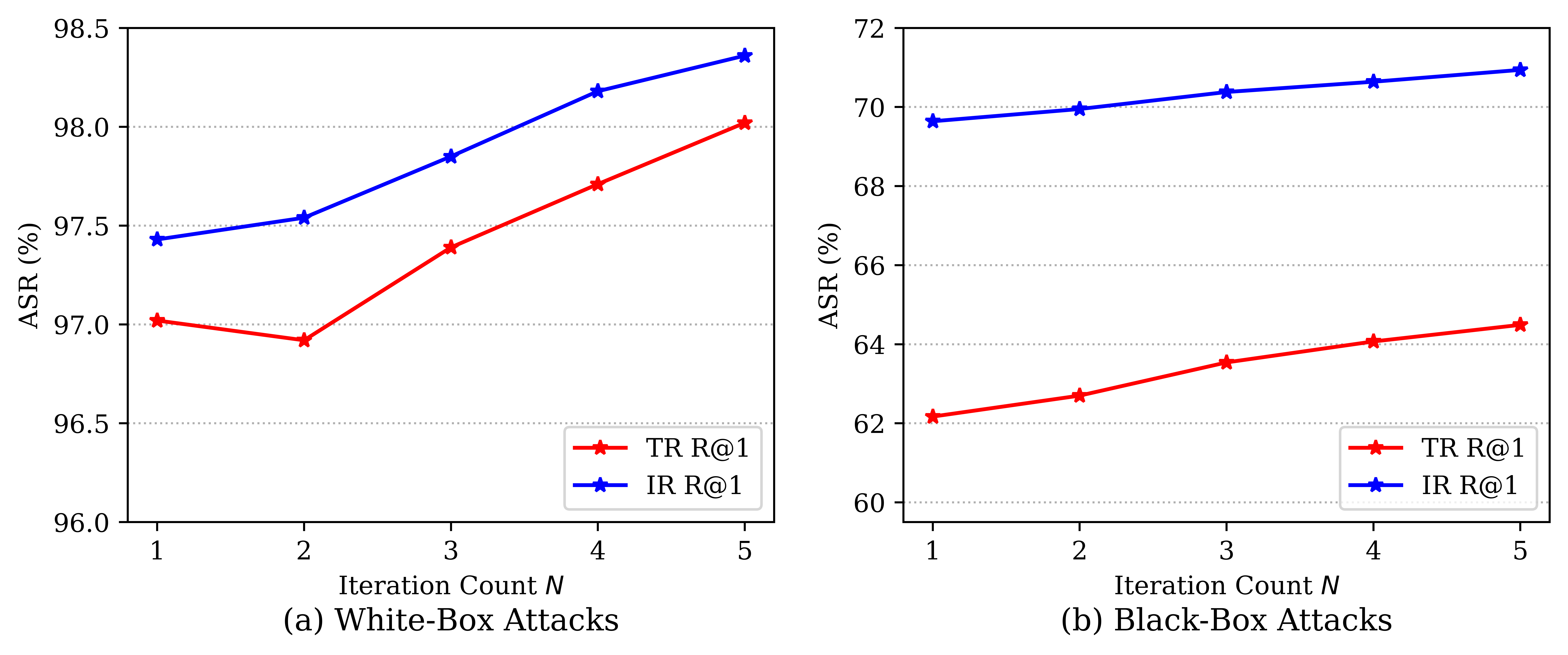} 
\caption{\textbf{Impact of Iteration Count $N$ on Interactive Enhancement.} We use the Flickr30K dataset, utilizing ALBEF as the source model and having both ALBEF and TCL as distinct target models. The trend indicates an overall improvement in ASR (\%) with the increasing $N$.}
\label{Figure 5}
\vspace{-8pt}
\end{figure}

\begin{figure*}[t]
\centering
\includegraphics[scale=0.71]{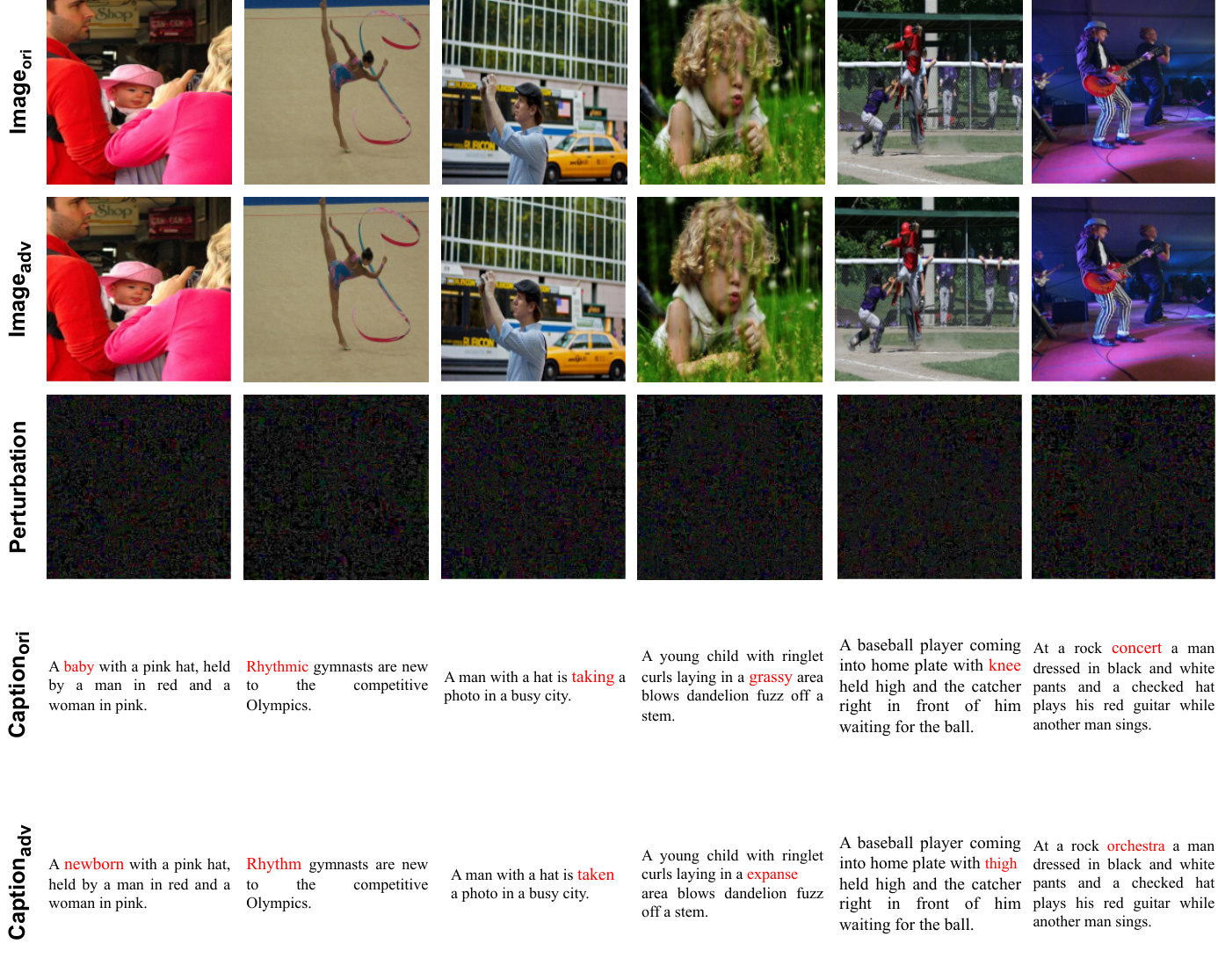} 
\caption{Visualization of CMI-Attack. The first row presents the original images, the second row displays adversarial examples generated by CMI-Attack, and the third row shows perturbations magnified 50 times for better visualization. The last two rows include the original caption and the caption after perturbation, respectively.}
\label{Figure 4}
\vspace{-8pt}
\end{figure*}

\noindent\textbf{Joint Gradient $\lambda$.}
To investigate the impact of different Joint Gradient values, we conduct tests with various $\lambda$ values, as shown in \cref{Table 4}. Excessively large joint gradients can lead to unstable convergence and increased difficulty in escaping poor local minima, while excessively small momentum coefficients may~make~the attacker more susceptible to noise interference from the input data, resulting in performance degradation. Overall, a relatively high success rate in transfer attacks is achieved when $\lambda=0.9$.

\noindent\textbf{Text Attack Strategies.}
For various text perturbation methods, we perform ablation experiments using WordNet, MLM, GloVe, and our combined GloVe and MLM approach. As indicated in \cref{Table 5}, our method demonstrates a notable superiority with a higher attack success rate in transfer attacks. It is noteworthy that when utilizing adversarial examples generated on ALBEF to attack the TCL model, our text attack strategy demonstrates a 13.49\% increase in the TR R@1 metric compared to using the MLM rule. This effectiveness is attributed to our dedicated attention to comprehending the influence of images on text within the multimodal alignment space during the text attack process, along with the strategic incorporation of embedding guidance.\par
\noindent\textbf{Iteration Count $N$ on Interactive Enhancement.} Similarly, for the number of interactive iterations $N$, we also conduct ablation analysis. Here, different count of $N$ represent the number of interactions between image gradient information and text embeddings during the perturbation of textual content. As shown in \cref{Figure 5}, as the value of $N$ increases, the instances of mutual interference between modalities also escalate, consequently leading to a discernible upward trend in the success rate of transfer attacks. In terms of white-box attacks, setting the iteration count $N$ to 5 results in higher TR R@1 and IR R@1 metrics by 0.94\% and 0.93\%, respectively, compared to when $N=1$. In the context of black-box attacks, under the same experimental configuration, there is an increase of 2.32\% and 1.10\%.
However, balancing performance and efficiency, we ultimately use $N=1$ in all experimental settings.

\subsection{Visualization}
In this section, we showcase the visualization results of CMI-Attack, as depicted in \cref{Figure 4}. The top row shows the original image, the second row depicts the image after perturbation, and the third row visualizes the perturbation effect amplified by 50 times. Even under 50 times amplification, the added perturbations are challenging to perceive, indicating the high imperceptibility of our attack. This observation underscores the effectiveness of the Collaborative Multimodal Interaction strategy in generating subtle and imperceptible adversarial perturbations. The fourth row contains the original caption associated with the image, while the fifth row displays the caption after the application of the adversarial attack. 
From the visualizations, it can be clearly seen that the perturbations of adversarial examples optimized with Embedding Guidance strategy are more imperceptible, validating the efficacy of this strategy.
Through these visualizations, we aim to provide~a~detailed and insightful portrayal of the effectiveness of CMI-Attack in generating imperceptible and impactful adversarial examples across multiple modalities.

\section{Conclusions}
\label{sec:conclusions}
In this paper, we conduct the first study on adversarial attacks against VLP models from the perspective of modality interaction, and propose a novel attack called CMI-Attack. 
CMI-Attack ingeniously leverages the intricate relationships between modalities by attacking text at the embedding level and enhancing constraints on perturbations of texts and images through the use of interaction image gradients, resulting in a significant improvement in the effectiveness of our adversarial attack.
We hope that this work can provide refreshing insights to advance an in-depth understanding of VLP robustness and promote safer model development.

\noindent\textbf{Broader impacts.}
Our work indicates that downstream tasks of VLP models are currently exposed to security risks. CMI-Attack aids researchers in understanding VLP models from the perspective of adversarial attacks, thereby facilitating the design of more reliable, robust, and secure VLP models.

\section*{Acknowledgments}
This work was supported by National Natural Science Foundation of China (No.62072112), Scientific and Technological innovation action plan of Shanghai Science and Technology Committee (No.22511102202), Fudan Double First-class Construction Fund (No. XM03211178).

\bibliographystyle{IEEEtran}
\bibliography{ijcai23}
\end{document}